%% file: paper.tex
\documentclass[conference]{IEEEtran}
\IEEEoverridecommandlockouts
\usepackage{adjustbox}
\usepackage{cite}
\usepackage{amsmath,amssymb,amsfonts}
\usepackage{graphicx}
\usepackage{textcomp}
\usepackage{xcolor}
\def\BibTeX{{\rm B\kern-.05em{\sc i\kern-.025em b}\kern-.08em
    T\kern-.1667em\lower.7ex\hbox{E}\kern-.125emX}}

\usepackage[ruled,vlined,linesnumbered]{algorithm2e}
\DontPrintSemicolon
\SetAlFnt{\small}

\SetCommentSty{algocommentstyle}
\usepackage{graphicx}
\usepackage{algpseudocode}
\usepackage{tikz}
\usetikzlibrary{calc}
\usepackage{listings}
\usepackage{float} 
\newfloat{lstfloat}{htbp}{lop}
\floatname{lstfloat}{Listing}
\usepackage{textcomp}
\usepackage{mdframed}
\usepackage{fancybox}
\usepackage{xspace}
\usepackage{upgreek}
\usepackage{graphicx}

\usepackage{hyperref}
\usepackage{soul}
\usepackage{booktabs}
\usepackage{comment}
\usepackage{cleveref}
\usepackage{ifthen}
\usepackage{enumitem}
\usepackage{soul}
\usepackage{multirow}
\usepackage{rotating}

\usepackage{makecell}
\usepackage{balance}
\usepackage{fancyvrb}
\usepackage{tcolorbox}
\usepackage{lipsum}  

\usepackage{tikz}
\usepackage{pgfplots}
\pgfplotsset{compat=1.15}
\usepgfplotslibrary{statistics}
\usetikzlibrary{fadings}

\usepackage{xcolor,pifont}
\newcommand*\colourcheck[1]{%
  \expandafter\newcommand\csname #1check\endcsname{\textcolor{#1}{\ding{52}}\xspace}%
}
\newcommand*\colourcross[1]{%
  \expandafter\newcommand\csname #1cross\endcsname{\textcolor{#1}{\ding{56}}\xspace}%
}

\newcommand{\head}[1]{\noindent\textbf{#1.}}

\begin{document}

\pagenumbering{arabic} 
\pagestyle{plain}

\title{Benchmarking Generative AI Models for Deep Learning Test Input Generation\\
\thanks{This work was supported by the project SOP CUP N.H73C22000890001, PNRR M4 C2 I1.3 “SEcurity and RIghts in the CyberSpace (SERICS)” PE0000014 PE7, funded by Next-Generation EU, and the Bavarian Ministry of Economic Affairs, Regional Development and Energy.}}

\author{\IEEEauthorblockN{Maryam}
\IEEEauthorblockA{
\textit{University of Udine} \\
Udine, Italy \\
maryam@spes.uniud.it}
\and
\IEEEauthorblockN{Matteo Biagiola}
\IEEEauthorblockA{
\textit{Universit\`a della Svizzera italiana} \\
Lugano, Switzerland \\
matteo.biagiola@usi.ch}
\and
\IEEEauthorblockN{Andrea Stocco}
\IEEEauthorblockA{
\textit{Technical University of Munich, fortiss} \\
Munich, Germany \\
andrea.stocco@tum.de}
\and 
\IEEEauthorblockN{Vincenzo Riccio}
\IEEEauthorblockA{
\textit{University of Udine} \\
Udine, Italy \\
vincenzo.riccio@uniud.it}
}

\maketitle

\input{0-abstract}
\begin{IEEEkeywords}
Software Testing, Generative AI, Deep Learning
\end{IEEEkeywords}

\maketitle

\input{1-introduction}
\input{2-background}
\input{3-test-generation-technique}
\input{4-empirical-setup}
\input{5-results}
\input{6-discussion}
\input{7-related-work}
\input{8-conclusion}

\balance
\bibliographystyle{IEEEtran}
\bibliography{paper}

\end{document}

%% file: 0-abstract.tex
\begin{abstract}
Test Input Generators (TIGs) are crucial to assess the ability of Deep Learning (DL) image classifiers to provide correct predictions for inputs beyond their training and test sets. 
Recent advancements in Generative AI (GenAI) models have made them a powerful tool for creating and manipulating synthetic images, although these advancements also imply increased complexity and resource demands for training.

In this work, we benchmark and combine different GenAI models with TIGs, assessing their effectiveness, efficiency, and quality of the generated test images, in terms of domain validity and label preservation.
We conduct an empirical study involving three different GenAI architectures (VAEs, GANs, Diffusion Models), five classification tasks of increasing complexity, and 364 human evaluations. Our results show that simpler architectures, such as VAEs, are sufficient for less complex datasets like MNIST. However, when dealing with feature-rich datasets, such as ImageNet, more sophisticated architectures like Diffusion Models achieve superior performance by generating a higher number of valid, misclassification-inducing inputs.
\end{abstract}

%% file: 1-introduction.tex
\section{Introduction}\label{sec:introduction}

Deep Learning (DL) has reshaped several fields, including image processing, where DL image classifiers often outperform traditional vision methods and even human experts in accuracy and efficiency~\cite{wang2021comparative}.
This advancement has enabled greater automation, especially in life- and safety-critical areas such as healthcare and autonomous driving~\cite{sarwinda2021deep, lou2022testing, tang2023survey, neelofar2024identifying}. 

Ensuring the quality of DL image classifiers remains crucial, as it is difficult to assess their ability to generalize to unseen data. In fact, their training and test sets may not fully capture the range of real-world scenarios they will encounter after deployment~\cite{guerriero2021operation,zohdinasab2023deepatash}.
A significant challenge for software testers is generating test images that accurately reflect real-world operating conditions and trigger \textit{misclassifications}, i.e., unexpected behaviors where predicted labels deviate from the expected ones. 
Thus, researchers have proposed test input generators (TIGs), aimed at automatically generating synthetic images to assess the quality of DL classifiers~\cite{zhangTSE22,RiccioEMSE20, braiek2020testing, 2023-Riccio-ICSE}. 

\begin{figure}[t]
\centering
\includegraphics[width=0.65\linewidth]{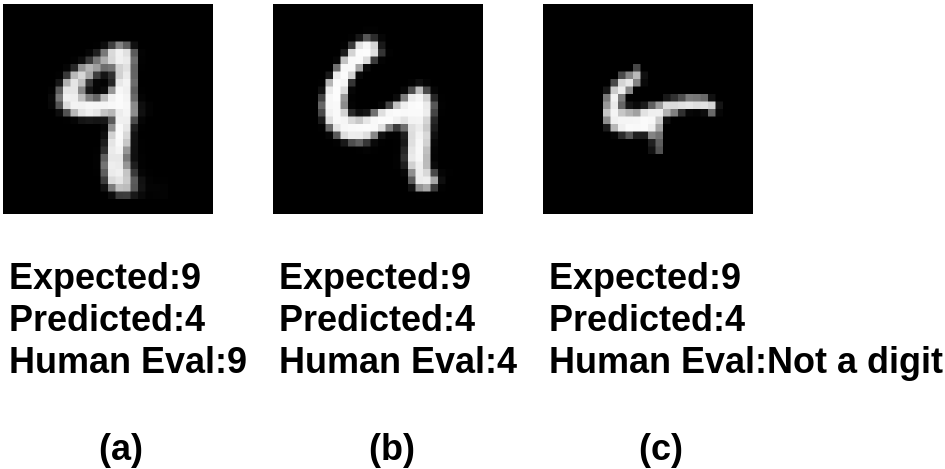} 
\caption{Misclassification-inducing test images for handwritten digit classifiers: (a) valid and label-preserving, (b) valid but not label-preserving, (c) invalid.}
\label{fig:intro}
\end{figure}

Most TIGs apply small perturbations to inputs to maintain their expected label, addressing the oracle problem~\cite{baresi2001test, pezze2014automated, jahangirova2016test}, as the correct output is unknown for synthetic data.
However, such perturbations can lead to invalid inputs (i.e., outside the classification task's domain) or fail to preserve the label of the original image (i.e., no longer belong to the same class)~\cite{dola2021distribution, 2023-Riccio-ICSE}.
Existing TIGs often overlook input validity and assume that ground-truth labels remain preserved~\cite{2023-Riccio-ICSE}.
\autoref{fig:intro} shows three misclassified inputs for an handwritten digit classifier: input (a) is valid and the expected label (i.e., 9) matches with the human assessment; (b) is valid but fails to preserve the expected label; and (c) is invalid, because it does not belong to the domain of handwritten digits.

The first propositions of TIGs primarily focused on manipulating raw inputs (i.e., pixel perturbations~\cite{pei2017deepxplore, guo2018dlfuzz, ma2018deepgauge, braiek2019deepevolution}) or parametrized semantic representations (e.g., control points in vector graphics~\cite{riccio2020model, zohdinasab2021deephyperion, zohdinasab2023deepatash} or parameters in simulators~\cite{Abdessalem-ICSE18, fahmy2021supporting}).
However, these approaches are restricted to modifying initial images with known ground-truth labels, limiting exploration to regions near the original inputs and leaving significant portions of the input space untested.

Researchers have recently started leveraging the creativity of distribution-aware Generative AI (GenAI) models~\cite{kang2020sinvad, dunn2021exposing, dola2021distribution, aleti2023software, dola2024cit4dnn}, which learn the input data distribution in the form of a \textit{latent space}, i.e., a compressed low-dimensional representation capturing the key features of the problem domain~\cite{goodfellow2020generative}. GenAI-based TIGs manipulate inputs in the latent space, where small changes can yield significant image variations (e.g., style, pose, color), before converting latent representations back into pixel space. In this way, GenAI models can generate unique images, blending features from learned patterns in ways not present in the original training data.

\begin{figure*}[t]
\centering
\includegraphics[width=\linewidth]{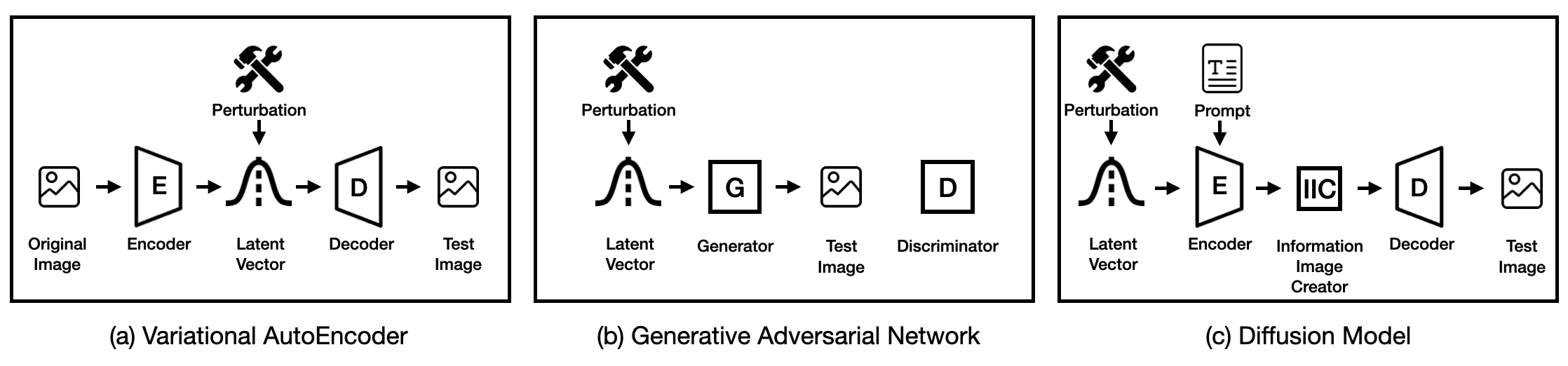} 
\caption{Summary of the GenAI models considered in this paper and the process by which TIGs perturb their latent vectors.}
\label{fig:GenAI}
\end{figure*}

Existing studies have demonstrated the effectiveness of GenAI-based TIGs~\cite{kang2020sinvad, dunn2021exposing, 2023-Riccio-ICSE, kang2024deceiving}. However, they are often influenced by confounding factors such as differences in testing algorithms and the absence of standardized training, experimental setups, and hyperparameter tuning. Moreover, they have not yet incorporated recent innovations like diffusion models, making it difficult to fairly compare the test generation capabilities across different GenAI models.

In our study, we fill this gap by providing the most comprehensive cross-evaluation of GenAI TIGs, benchmarking three key GenAI models, i.e., VAEs, GANs, and diffusion models, across popular datasets of increasing complexity. We provide a standardized experimental framework, including search-based optimization for latent space manipulation, and systematically evaluate these models in terms of their ability to generate valid, label-preserving, and misclassification-inducing inputs, considering efficiency and effectiveness.
Our results show that GenAI TIGs can produce valid, label-preserving inputs, although the degree of success varies. Simpler models, such as VAEs and GANs, perform well on less complex datasets like MNIST and SVHN. Diffusion models excel in complex tasks like CIFAR-10 and ImageNet, generating up to 80\% more valid label-preserving inputs. 
Our key contributions are:
\begin{itemize}
    \item A large-scale empirical study comparing three GenAI architectures (VAEs, GANs, and diffusion models) across five classification tasks, along with a validity assessment involving 364 human assessors. 
    \item A search-based test generation framework that integrates different GenAI models, enabling automated testing of DL classifiers through latent space manipulation.
\end{itemize}

To encourage open research, our test generation framework and experimental data are available~\cite{repo}.

%% file: 2-background.tex
\section{Background}\label{sec:background}

In the following, we describe the three main GenAI architectures considered in this paper (\autoref{fig:GenAI}).

\subsection{Variational Autoencoders (VAEs)} 
VAEs, introduced by Kingma et al.~\cite{kingma2013auto}, consist of two neural networks: an encoder and a decoder (\autoref{fig:GenAI}~(a)). The encoder learns to map images to lower-dimensional representations (i.e., latent space vectors sampled from a predefined distribution). The decoder learns to reconstruct images from the latent vectors. 
This architecture extends standard autoencoders by adopting a probabilistic approach to model complex data distributions. The two networks are trained jointly to minimize a loss function that balances two objectives: the reconstruction error (i.e., difference between original inputs and VAE's reconstructions), and the Kullback-Leibler divergence~\cite{10.1214/aoms/1177729694}, which regularizes the learned latent space distribution to align with a prior distribution, e.g., a standard normal distribution. The sampling step during training introduces randomness, enabling VAEs to learn smooth and continuous latent spaces.

In software testing, Kang et al.~\cite{kang2020sinvad, kang2024deceiving} and Dola et al.~\cite{dola2021distribution, dola2024cit4dnn} used VAEs for generating edge cases to assess the robustness of image classifiers. In particular, their TIGs encode existing images to increase control over generation and perturb the latent vectors returned by the encoder.

\subsection{Generative Adversarial Networks (GANs)} 
GANs introduced by GoodFellow et al.~\cite{goodfellow2020generative}, consist of two neural networks, i.e., a generator and a discriminator, trained simultaneously (\autoref{fig:GenAI} (b)). These networks are involved in competitive learning to improve one another performance: the discriminator learns to distinguish real from artificially generated images, while the generator learns to create images from latent vectors sampled from a known distribution that can \textit{fool} the discriminator.
While VAEs explicitly model the data distribution and enforce smoothness in the latent space trough their probabilistic approach, GANs' generators implicitly learn to approximate the data distribution by continuously improving their ability to deceive discriminators. This often results in more realistic images than those generated by VAEs, although GANs are more challenging to train due to issues like instability and mode collapse, where the generator learns only a limited subset of features, leading to reduced diversity in the generated images~\cite{goodfellow2020generative}.
GANs are useful for testing, as they can generate realistic artificial images that resemble the training distribution and are difficult to distinguish from real data. Moreover, GANs allow test manipulation by perturbing latent space vectors (i.e., input to the GAN generator).
In software testing, Dunn et al.~\cite{dunn2021exposing} adopt conditional Deep Convolutional GANs (cDCGANs), which leverage convolutional layers for improved image generation quality, and incorporate additional information (called \textit{conditions}) such as class labels to better guide the generation of images that meet predefined criteria.

\subsection{Diffusion Models (DMs)} 
DMs were first introduced by Sohl-Dickstein et al.~\cite{sohl2015deep}.
The core idea behind these models is to simulate a diffusion process, where information in an image is gradually diffused (i.e., noise is incrementally added) and then progressively recovered through a reverse process, known as denoising (\autoref{fig:GenAI} (c)). 
During training, diffusion models add noise to the data over a sequence of steps, gradually transforming an original vector into a noisy version that resembles a simple Gaussian distribution. The model is then trained to reverse this process and recover the original data, learning how to transform noise back into a complex data distribution, i.e., natural images~\cite{ho2020denoising}.
This approach enables the generation of highly realistic images, offering significant advantages over other generative models, such as GANs, particularly in terms of training stability and resistance to mode collapse~\cite{sohl2015deep}.

In text-to-image scenarios, we considered models like Stable Diffusion, which leverages a textual description of the desired input (called prompt) to generate corresponding images. Its architecture consists of three main components: the text encoder, the image information creator, and the image decoder. 
A noisy latent vector is obtained by combining a noise vector with the latent vector obtained by encoding the prompt. The information image creator iteratively refines this noisy latent vector, guided by the prompt. The final latent vector is then transformed into a high-resolution image by the decoder.
This work is the first to integrate Stable Diffusion into a TIG for image classifiers, by manipulating its input latent vector.

%% file: 3-test-generation-technique.tex
\section{Evaluation Framework}\label{sec:approach}

In this section, we describe the evaluation framework we propose for comparing  different GenAI architectures for the generation of misclassification-inducing inputs for DL based image classifiers. Our framework is designed to (1)~smoothly integrate different GenAI models, (2)~enable the generation and controlled manipulation of novel inputs through the corresponding latent vectors, (3)~guide input manipulation towards inducing misclassifications, and (4)~return the misclassification-inducing inputs along with statistics on the efficiency of their generation.
To explore the latent space produced by the GenAI methods, we adopt search-based optimization using a population-based genetic algorithm, due to its success in testing DL based systems~\cite{riccio2020model, kang2020sinvad,Abdessalem-ICSE18,Abdessalem-ASE18-1,Abdessalem-ASE18-2,gambi2019automatically}. 
Specifically, we define a fitness function that assigns lower values to inputs more likely to induce misclassifications, which the search process seeks to minimize.

\input{algorithm}

\autoref{alg:tig} outlines the steps of our framework. It takes as input the classifier under test $C$, the latent vector used by the GenAI model to produce an image (i.e., the \textit{seed}), the corresponding expected label (i.e., the class the image should belong to), and the configuration parameters for the test generation process. As output, it returns the first misclassification-inducing input it identifies, along with the number of iterations required to generate it.

The algorithm starts by initializing the perturbation step, which is used to modify the latent vectors of the GenAI model $G$ (Line~1).
The first image is generated from the seed latent vector, which provides the necessary information for $G$ to produce an image belonging to the target class (Line~2).
This first image is evaluated on the classifier under test, determining the predicted label and its fitness score (Line~3). 
The test generation process terminates if the initial image is misclassified, i.e., it does not belong to the expected class according to the classifier (Lines~4-5). In fact, the goal of a TIG is to find slight perturbations that transform an image predicted as expected into another one predicted differently, as these can highlight weaknesses of the classifier~\cite{riccio2020model}.
A population of size \textit{popSize} is created through the mutation genetic operator, which applies random perturbations to the initial latent vector, thus increasing diversity within the population.

The main loop is executed up to $N$ times (Lines~9--31), i.e., it stops if a misclassification is found or if the iteration budget is exhausted.
At each iteration, $G$ produces images corresponding to the latent vectors in the population $P$, which are evaluated by the classifier (Lines~10-11).
The selection operator chooses the most promising inputs, i.e., those with lower fitness (Line~12). If an individual exhibits negative fitness (indicating a misclassification), the corresponding image is returned as output and the algorithm terminates (Lines~14--16). Otherwise, the selected individuals are modified using the genetic operators.

The mutation operator is adaptive, i.e., the perturbation step increases if no improvement in fitness is observed during the previous iteration (Lines~18--21).
Offsprings are obtained by applying the mutation and crossover operators (Lines~23--29). To prevent the generation of unrealistic images, latent vectors are clamped within the bounds \textit{minBound} and \textit{maxBound} observed during the GenAI model training, as suggested by Dunn et al.~\cite{dunn2021exposing}.
The offsprings and the best individuals from the current iteration form the population for the next iteration (Line~30).
In the following, we describe in detail each main component of the search algorithm.

\subsection{Input Representation and Initialization} \label{sec:init_pop}

To represent the input domain, our framework leverages a lower-dimensional latent space synthetized by the GenAI models. 
The latent space captures underlying patterns of the input data in a more abstract and compressed form compared to the original high-dimensional input space.
It consists of a multivariate probability distribution characterized by parameters such as the means and variances of a predefined $z$-dimensional distribution chosen during training.

Our approach represents test inputs as latent vectors, which are points sampled within the latent space, whose dimensionality is defined by the specific GenAI model adopted by the TIG. Exploring the latent space is more efficient than searching directly in high-dimensional pixel space due to its reduced dimensionality and mapping to a known distribution.
Our TIG leverages an initial seed latent vector corresponding to an input image whose label is known and correctly predicted by the classifier under test. From this seed, our TIG produces a population of slight variations, enabling exploration of its neighborood and introducing diversity into the search process. The starting seed depends on each specific GenAI model.

VAEs use the latent vector produced by the encoder when processing an image from the original test set of the considered dataset, whose ground-truth label is known. This approach, used in the literature by Kang et al.~\cite{kang2020sinvad} and Dola et al.~\cite{dola2024cit4dnn}, ensures higher similarity to the original input and, thus, label preservation and better control over the exploration.

As for GANs, the initial seed is a vector sampled from the target distribution used during training. Specifically, we use \textit{conditional} GANs, which accept as input both the latent vector and the target label. This conditioning, also employed by Dunn et al.~\cite{dunn2021exposing}, improves control over image generation.

Also for DMs, the seed is a vector sampled from the training distribution. This GenAI model controls the generation by accepting as input also a textual prompt, which describes both the input domain and the target class.
Although no current TIG employs this approach, latent space exploration for DMs is a well-established practice in the field~\cite{SD_latent_walk}.
Stable Diffusion \textit{guidance scale} parameter significantly influences the image generation process. This parameter determines to what extent the generated image adheres to the prompt. Higher values make the model focus more on the prompt, resulting in images closely aligned with the input text, but may reduce diversity and occasionally lead to lower image quality. Lower values introduce more diversity and creativity, with a looser connection to the prompt.
During seed generation, we adopt a relatively higher value (3.5)
for the DMs guidance scale parameter compared to latent vector mutation (1.4). This choice, motivated by preliminary experiments, promotes label adherence during mutation and diversity in seed generation, as recommended by the authors' guidelines~\cite{nichol2022glidephotorealisticimagegeneration, getimg_ai_guide}.

\subsection{Fitness Function}

The \textsc{Evaluate} function calculates the fitness of each individual in the population, based on how the classifier under test predicts the image generated from the corresponding latent vector.
In this context, an individual is considered fit if it is likely to cause a misclassification, i.e., the corresponding image should belong to one class but is classified as a different class.
To this aim, we leverage the activation levels from the output softmax layer of the image classifier. It provides a surrogate confidence score for each possible class~\cite{pmlr-v70-guo17a} as the softmax output can be interpreted as a probability distribution over the classes, with the highest value indicating the predicted class for a given input.
Similar fitness functions have been employed by several existing approaches~\cite{pei2017deepxplore, guo2018dlfuzz, riccio2020model, dunn2021exposing}.

Specifically, the fitness value is computed as the difference between the confidence score of the expected class and the highest confidence score for any other class. The formula of the fitness for the input image $x$ is defined as follows:

\begin{equation}
   \textit{fitness}(x) = \sigma_{\text{expected}}(x) - \max_{i \neq \text{expected}} \sigma_i(x) 
\end{equation}

\noindent
where $\sigma_{\text{expected}}(x)$ represents the softmax output for the expected class, and $\max_{i \neq \text{expected}} \sigma_i(x)$ is the highest softmax output for any other class. The fitness function approaches zero when the confidence scores of the expected class and the second-highest class are similar. A negative value indicates a misclassification, as the highest confidence score is assigned to a class different from the expected one. Thus, this fitness function is designed to be minimized by the test generator to promote inputs that are closer to induce misclassifications.

\subsection{Genetic Operators}
Our \textsc{Select} operator chooses the most promising individuals for the next generation, i.e., those with the lowest fitness scores. The number of individuals chosen is controlled by the parameter \textit{tshdBest}, which can be tuned to balance exploration and exploitation. A lower \textit{tshdBest} value intensifies exploitation by focusing on fewer, high-performing individuals, potentially speeding up failure detection but with the risk of premature convergence to local optima; a higher value, instead, retains more individuals, promoting greater diversity in the population and allowing for a broader exploration.

Representing inputs as latent vectors enables uniform \textsc{Mutation} and \textsc{Crossover} operators across all GenAI models.
We adopt a single-point \textsc{Crossover} operator that combines two individuals randomly chosen among the ones selected by the selection operator. This operator divides the latent vectors of the parents at a random point, taking the first portion from one parent and the second portion from the other, generating an offspring that inherits characteristics from both. This promotes exploration of new latent space regions by combining information from the most successful individuals.

The \textsc{Mutation} operator introduces random perturbations to the offspring, encouraging diversity and preventing stagnation in local optima. These perturbations are obtained by multiplying random noise sampled from a normal distribution by the perturbation step, as described in the following formula: 

\begin{equation}
   \mathbf{z}_{mut} = \mathbf{z}_{orig} + \boldsymbol{\epsilon} \cdot \delta, \quad \boldsymbol{\epsilon} \sim \mathcal{N}(0, 1)^{d}
\end{equation}

\noindent
where $z_{orig}$ is the initial latent vector of dimensionality $d$, $\epsilon$ is the noise vector of the same dimensionality where each component is sampled from a standard normal distribution, $\delta$ is the perturbation step, and $z_{mut}$ is the perturbed latent vector.

The perturbation step $\delta$ is adjusted adaptively based on the fitness of the previous iteration, i.e., if the previous iteration did not improve the best fitness score, then the mutation extent is increased to escape local optima by allowing larger changes to the latent vectors. Conversely, if improvement is observed, the mutation rate is decreased to its default value $\delta_{init}$, in order to exploit promising areas of the solution space.

%% file: algorithm.tex
\SetKwComment{Comment}{/* }{ */}
\begin{algorithm}[t]

\caption{Test Generation with GenAI Models}
\label{alg:tig}

\footnotesize

\KwData{
$s$: seed latent vector, 
$\textit{expLabel}$ expected label, 
$G$: GenDL model, 
$C$: classifier under test, 
$\textit{popSize}$: population size, 
$\textit{tshdBest}$: threshold for selection operator, 
$N$ number of iterations, 
$\delta_{\textit{init}}$: initial perturbation step, 
$\textit{minBound}$, 
$\textit{maxBound}$: minimum and maximum values observed during the training of $G$.}
\KwResult{
$\hat{\textit{img}}$: misclassification-inducing image, 
$\textit{iter}$: \# of iterations needed to generate a test input}

perturbation step $\delta \gets \delta_{\textit{init}}$\; 
\text{image} $\textit{img} \gets G(s)$\;
$\text{label } l, \text{fitness } f_{\textit{prev}} \gets \textsc{Evaluate}(\textit{img}, C, \textit{expLabel})$\;
\If{$(l \neq \textit{expLabel})$}{
    \Return {$\emptyset$, $0$} \Comment{Return if the original input is misclassified}
}

\For{$i = 1 \to \textit{popSize}$}{
population individual $P_i \gets \textsc{Mutation}(s, \delta)$\;
}
$\textit{iter} \gets 0$\;
\While{$\textit{iter} < N$}{
    \text{image population} $I_P \gets G(P)$\;
    $\text{fitness values } F \gets \textsc{Evaluate}(I_P, C, \textit{expLabel})$\;
    \text{best individuals } $P' \gets \textsc{Select}(P, F, \textit{tshdBest})$\;
    \text{best fitness }$f_{\textit{min}} \gets \min(F)$\;
    \If{$(f_{\textit{min}} < 0)$}{
    $\hat{\textit{img}} \gets \textsc{getBestIndividual}(P')$\;
    \Return{$\hat{\textit{img}}$, $\textit{iter}$}}\;
\ElseIf{$(f_{\textit{min}} == f_{\textit{prev}})$}{
    $\delta \gets 2 \cdot \delta$\; \Comment{Adaptively increase the mutation extent}
}
\Else{
    $\delta \gets \delta_{\textit{init}}$\;  \Comment{reset to initial mutation extent when fitness improves}
    $f_{\textit{prev}} \gets f_{\textit{min}}$
}
    \text{offspring} $O \gets [ ]$\;
    \For{$j \gets 1$ \KwTo $(\textit{popSize} - \textit{length}(P'))$}{
        \text{parents} $p_{j1}, p_{j2} \gets \textsc{RandomChoice}(P', 2)$\;
        \text{offspring} $o \gets \textsc{Crossover}(p_{j1}, p_{j2})$\;
        $o \gets \textsc{Mutation}(o, \delta)$\;
        $o \gets \textsc{Clamp}(o, \textit{minBound}, \textit{maxBound})$\;
        $O \gets O \cup \{o\}$\;
    }
    $P \gets P' \cup O$\; 
    $\textit{iter} \gets \textit{iter} + 1$\;
    }
    \Return{$\emptyset$, $\textit{iter}$}\;

\end{algorithm}

%% file: 4-empirical-setup.tex
\section{Empirical Setup}\label{sec:empirical-study}

Our study compares different GenAI models for test input generation. We integrate each of them into our framework (\autoref{sec:approach}) to create \textit{GenAI TIGs}. We assessed their effectiveness and efficiency in generating valid, label-preserving inputs that cause misclassifications in the classifier under test.

\subsection{Research Questions}

\textbf{RQ\textsubscript{1} (Seed Generation):} \textit{Which GenAI model generates more correctly classified seed inputs within the same budget?}

Effective test generation begins with the identification of reliable seeds, i.e., inputs that produce images that are predicted as expected by the classifier under test~\cite{riccio2020model}. In fact, only these seeds can be used by TIGs to identify input variations that trigger misclassifications. 

\textit{Metric:} For each GenAI model, we compute the ratio of seeds assigned the correct label by the classifier, compared to the total number of generated seeds.

\textbf{RQ\textsubscript{2} (Effectiveness):} \textit{Which GenAI TIG generates more misclassification-inducing inputs?}

A standard approach to evaluate and compare the effectiveness of TIGs is to count the number of failures triggered within a given budget~\cite{zhangTSE22}. A TIG that triggers more misclassifications could potentially expose more weaknesses or bugs in the classifier under test.

\textit{Metric:} For each TIG, we calculate the number of misclassification-inducing inputs and their ratio over the total number of generated inputs.

\textbf{RQ\textsubscript{3} (Efficiency):} \textit{How efficient are GenAI TIGs in triggering misclassifications?}

The goal of this research question is to assess the efficiency of each GenAI TIG in triggering misclassifications during the iterative process of our testing framework. Efficiency, in this context, refers to the model's ability to generate misclassification-inducing inputs with fewer iterations, providing insight into how quickly each TIG can expose weaknesses in the classifier under test.

\textit{Metric:} We measure the average number of iterations required to trigger a misclassification across all seeds. For seeds where the TIG does not trigger any misclassification, we report the maximum number of iterations, i.e., the search budget.

\textbf{RQ\textsubscript{4} (Validity):} \textit{Which GenAI TIG generates more valid, misclassification-inducing inputs, according to humans?}

TIGs offer a reliable assessment of classifiers' quality only when misclassification-inducing inputs are valid, i.e., recognisable by humans as part of the input domain~\cite{2023-Riccio-ICSE, zhang2024enhancing}. Invalid inputs may not be worth further analysis by testers, as they refer to images that cannot be observed in the real world and, thus, do not provide meaningful insights into the defects of the DL model under test. In this work, we evaluated each generated input with multiple independent human assessors.

\textit{Metric:} We calculate the number and ratio of valid inputs (as assessed by human evaluators) over the total number of misclassification-inducing inputs.

\textbf{RQ\textsubscript{5} (Label Preservation)}
\textit{To what extent the valid misclassification-inducing inputs generated by GenAI TIGs preserve the seed's label?}

A misclassification is detected when the predicted label differs from the expected one, i.e., the seed's label. For this reason, it is crucial to evaluate whether the images generated by TIGs preserve the expected label.

\textit{Metric:} For each TIG, we measure the number and ratio of valid, misclassification-inducing, and label-preserving inputs over the total number of generated, valid and misclassification-inducing inputs.

\subsection{Datasets and Image Classifiers} 

\begin{table}[t]
\caption{Datasets' input size and classifiers' training accuracy.}
\label{tab:classifiers}
\begin{adjustbox}{width=\columnwidth}
\begin{tabular}{lrrrr}
\toprule
\bf{Dataset} & \bf{Image Size} & \bf{Classifier} & \bf{Accuracy (\%)} \\ 
\midrule 
MNIST & 28x28x1 & deepconv~\cite{kang2020sinvad} & 99.46 \\  
SVHN & 32x32x3 & VGGNET~\cite{kang2020sinvad} & 95.20 \\ 
CIFAR-10 & 32x32x3 & VGGNET~\cite{kang2020sinvad} & 86.38 \\ 
ImageNet & 224x224x3 & VGG19bn~\cite{pytorch-vgg-source} & 75.00\\ 
\bottomrule 
\end{tabular}
\end{adjustbox}
\end{table}

We consider four widely-used image datasets of increasing complexity: MNIST~\cite{LecunBBH98}, SVHN~\cite{Netzer2011}, CIFAR-10~\cite{Krizhevsky2009}, and ImageNet~\cite{Deng2009}. These datasets cover classification tasks ranging from recognizing 10 classes in small greyscale images to identifying one of 1,000 classes from large-scale, colored images representing real-world objects. In the following, we describe the considered datasets and classifiers (\autoref{tab:classifiers}).

\head{MNIST} 
A dataset of 70,000 grayscale images of handwritten digits, each with a resolution of 28x28 pixels and pixel values ranging from 0 to 255. The labels correspond to digits from 0 to 9. Due to its simplicity, MNIST is widely used in DL frameworks' tutorials~\cite{tensorflow-keras-example} and research papers~\cite{RiccioEMSE20}. The classifier under test is the convolutional DNN used in the TIG proposed by Kang et al.~\cite{kang2020sinvad}, which includes four convolutional layers, two pooling layers, and two fully connected layers.

\head{SVHN} 
This dataset contains 600,000 images of house numbers, representing digits from 0 to 9. Unlike MNIST, this dataset poses a more difficult challenge, with larger, colored images (32x32), where the target digit may be surrounded by neighboring digits, adding complexity to the recognition task. The classifier under test is the VGG architecture used by Kang et al.~\cite{kang2020sinvad}, which employs five convolutional blocks.

\head{CIFAR-10} 
This dataset consists of 60,000 color images, each 32x32 pixels, spanning 10 different classes representing animals and vehicles. This dataset is more complex than SVHN because it involves a wider variety of classes with greater diversity and complexity in terms of backgrounds, textures, lighting, and object orientations. For this dataset, we used the same DL architecture used for SVHN~\cite{kang2020sinvad}.

\head{ImageNet-1k} dataset consists of over 14 million images spanning 1,000 classes. It is widely recognized for its role in the ImageNet Large Scale Visual Recognition Challenge (ILSVRC) since 2010, with the 2012 version being a benchmark standard for image classification tasks. Compared to the other three datasets, ImageNet-1k includes high-resolution images and a significantly broader range of categories. We tested the VGG19bn pretrained DL model, available through the PyTorch library~\cite{pytorch-vgg-source}, consisting of 16 convolutional layers and three fully connected layers.

\subsection{GenAI Models Setup}

We compared the GenAI models introduced in \autoref{sec:background}. To ensure fairness, each model was trained on the same training set originally used to train the DL classifier under test. The latent space of these GenAI models was leveraged in combination with the TIG described in \autoref{sec:approach} to generate test images. \autoref{tab:gendl} provides an overview of each GenAI model, including a link to its architecture and details about its training and inference process, offering insights into their respective computational costs and complexities. The models were trained until convergence, on a machine featuring an NVIDIA GeForce RTX 3090 GPU (24 GB VRAM) and CUDA 12.4 for GPU
acceleration.

\input{genai}

\head{VAE} 
We trained unconditional VAEs for all the considered datasets. In particular, we trained increasingly more complex architectures, aligning with the difficulty of the classification task. They span from a basic architecture with a single fully connected hidden layer in both the encoder and decoder for MNIST\cite{kang2020VAEMNISTcode}, to more sophisticated architectures, incorporating multiple fully connected and convolutional layers to effectively capture the features of ImageNet data \cite{maunish2023VAEImagenetkaggle}.

\head{GAN} 
For MNIST, SVHN and CIFAR-10, we adopted a flexible conditional deep convolutional GAN from the literature~\cite{luo2021case} and trained it on each dataset separately. As dataset complexity increases, it becomes more challenging for the generator to produce realistic images. For this reason, we adopted for ImageNet the more sophisticated conditional BigGAN architecture~\cite{brock2018large} used by Dunn et al.~\cite{dunn2021exposing}.

\head{DM} 
Due to the high complexity and computational demands of the Stable Diffusion model, it was not feasible to train it from scratch on each dataset. 
For this reason, we fine-tuned the pretrained, robust Stable Diffusion model v1.5~\cite{stable-diffusion-v15} provided by HuggingFace for five epochs on each dataset. We employed the Low-Rank Adaptation (LoRA) technique \cite{hu2021lora}, which reduces memory consumption and accelerates fine-tuning of large models. For each dataset, we defined specific textual prompts to guide the model in learning the concept of each class. Stable Diffusion was fine-tuned on the entire MNIST, SVHN, and CIFAR-10 datasets. Due to the large size and number of classes of ImageNet, we focused on fine-tuning two specific classes separately, i.e., \textit{teddy bear} and \textit{pizza}. 

\subsection{Experimental Procedure}

 We conducted a comprehensive comparison of the considered GenAI models, i.e., VAEs, GANs, and DMs, within our TIG framework. 
Each model was trained on five tasks across four datasets, resulting in a total of 15 distinct test generators.

For each GenAI model, we generated 100 starting seeds, according to the specific initialization for each model, as described in \autoref{sec:init_pop}.
Each TIG was allocated a budget of 250 iterations with consistent genetic algorithm parameters, i.e., a population size \textit{popSize} of 25 and a selection threshold \textit{tshdBest} of 10, maintaining the same ratio as related work~\cite{kang2020sinvad}. \autoref{fig:1} shows examples of misclassification-inducing images generated by our GenAI TIGs.
We experimented with two initial perturbation steps $\delta_{\textit{init}}$ that we chose as partitions of latent vector ranges of each GenAI model. After generating 1,000 seeds per model, we computed the range based on the maximum and minimum values of the latent vectors. In this way, we accounted for each GenAI model's unique latent space structure. The lower perturbation step ($\delta_{\textit{init}} = \textit{Low}$) was obtained by dividing the range by $10^{4}$, while the higher ($\delta_{\textit{init}} = \textit{High}$) by $10^{3}$. This allowed us to compare the impact of fine-grained vs more substantial perturbations on test generation effectiveness and efficiency.

The validity and label preservation of misclassification-inducing inputs were assessed by independent human evaluators.
Automated distribution-based validators~\cite{dola2021distribution,stocco2020misbehaviour} were not considered, as they are limited to checking whether inputs are in the same distribution of their training data and overlook the problem of label preservation, as shown by Riccio and Tonella~\cite{2023-Riccio-ICSE}. 
Instead, humans can provide more accurate judgments on whether the generated images semantically belong to the intended input domain.
We used the Amazon Mechanical Turk~\cite{mturk} crowdsourcing platform, which grants access to a diverse and independent group of assessors~\cite{behrend2011viability}. 
This platform is well-suited for qualitative feedback collection, as widely demonstrated in previous studies~\cite{KitturCS08,HeerB10,riccio2020model,2023-Riccio-ICSE}. Assessors were compensated appropriately~\cite{PastoreMF13}.

For each image, we asked the assessors to identify which class was represented within the problem domain or if the image did not belong to that domain (i.e., it was invalid). For MNIST, SVHN, and CIFAR-10, the assessors selected from the $10$ possible classes or the invalid option (``Not a handwritten digit/house number/real-world object''). 
Since ImageNet-1K has 1,000 classes, the user could choose between the expected class, the eight most commonly predicted classes by our classifier under test for the corresponding tasks, or the ``Another real-world object'' and ``No real-world objects'' options.
We implemented two quality control measures to assess the reliability of the responses provided by human assessors. First, we added an Attention Check Question (ACQ) to each survey. Second, we restricted the participation to workers with high reputation, i.e., above 95\% approval rate~\cite{peer2014reputation}. The ACQ consisted of an image for which the human choice is obvious, and only users passing this check were included in the results.

Surveys consisted of multiple questions for each classification problem, ensuring that each image was shown only once across the surveys.
Each survey was answered by $2$ assessors, so that each input was assessed by $2$ human evaluators, with a total of $182$ surveys and $364$ human participants.
We counted the number of images where both assessors agreed on the validity (i.e., assigned a class within the problem domain) or invalidity. Disagreements were excluded from the analysis. We then compared metrics for each pair of GenAI TIGs within the same classification task and perturbation step.

To determine statistical significance, we performed Fisher's exact test~\cite{Crowdsourcing6} for binary variables. Since the average number of iterations is a continuous variable, we adopted the Mann-Withney U-Test~\cite{Wilcoxon1945} when assessing efficiency, measuring the magnitude of the differences with the Cohen's D~\cite{cohen1988statistical}.
We threshold the $p$-value to be lower than $0.05$, combined with a non-negligible effect size or odds ratio, to assess a statistically significant difference between the compared GenAI TIGs.

%% file: genai.tex
\begin{table}[t]
\caption{Characteristics of the GenAI models: latent vector size, training time until convergence, average inference time.}
\label{tab:gendl}
\centering
\begin{adjustbox}{width=\columnwidth}
\begin{tabular}{llrrr}
\toprule
\textbf{Dataset} & \textbf{Model} & \textbf{LV size} & \textbf{$t_{train}$ (min)} & \textbf{$t_{infer}$ (ms)} \\ \midrule

\multirow{3}{*}{MNIST} 
& VAE \cite{kang2020VAEMNISTcode} & 400 & 6 & 0.27 \\
& GAN \cite{pytorch_dcganCode},\cite{pytorch_cdcganarchitecture} & 100 & 9 & 0.7 \\
& DM \cite{kohya_ss_lora_gui_codefile} & 16384 & 405 & 960.68 \\ 

\midrule

\multirow{3}{*}{SVHN} 
& VAE \cite{sinvad2020VAEConvCodearchitecture} & 800 & 93 & 4.07 \\
& GAN \cite{pytorch_dcganCode},\cite{pytorch_cdcganarchitecture} & 100 & 86 & 1.75 \\
& DM \cite{kohya_ss_lora_gui_codefile} & 16384 & 572 & 1213.49 \\ 

\midrule

\multirow{3}{*}{CIFAR-10}
& VAE \cite{sinvad2020VAEConvCodearchitecture} & 1024 & 423 & 2.51 \\
& GAN \cite{pytorch_dcganCode},\cite{pytorch_cdcganarchitecture} & 100 & 450 & 1.73 \\
& DM \cite{kohya_ss_lora_gui_codefile} & 16384 & 362 & 1903.29 \\ 

\midrule

\multirow{3}{*}{ImageNet}
& VAE \cite{maunish2023VAEImagenetkaggle} & 512 & 2521 & 11.92 \\
& GAN \cite{brock2018large} & 128 & 21600 & 20.68 \\
& DM \cite{kohya_ss_lora_gui_codefile} & 16384 & 30 & 1945.77 \\

\bottomrule

\end{tabular}%
\end{adjustbox}
\end{table}

%% file: 5-results.tex
\section{Results}\label{sec:results}

\begin{figure}[t]
\centering
\includegraphics[width=0.91\linewidth]{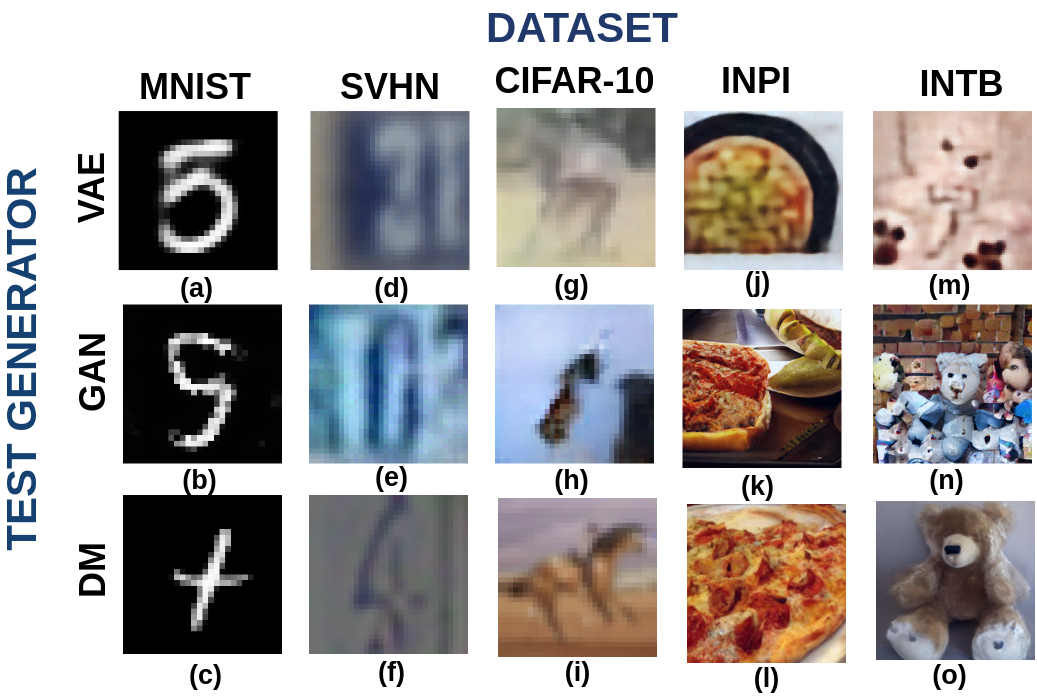} \caption{Misclassification-inducing images generated by GenAI TIGs} \label{fig:1}
\end{figure}

\input{table_for_Compare_model}

\subsection{RQ$_1$ (Seed Generation)}

The fourth column of \autoref{tab:comparison} reports the number of seeds generated by each GenAI model that were correctly predicted by the classifiers under test. The values are consistent across both low and high perturbation steps (``Low'' and ``High''), as we used the same initial seeds for each GenAI TIG.

For MNIST, both VAE and GAN produced a significantly higher percentage of correctly classified seeds compared to DM. This may be due to the fact that the original DM was trained on colored images, and during fine-tuning, it had to adapt to greyscale inputs.

For SVHN and CIFAR-10, however, DMs significantly outperformed VAEs and GANs, demonstrating better adaptability to the complexity of these datasets.

In the ImageNet tasks, GANs performed best for the \textit{pizza} class and similarly to DMs for the \textit{teddy bear} class. This improvement in GAN performance is probably due to the usage of the larger BigGAN architecture~\cite{brock2018large}, which is better suited for high-complexity datasets.

\begin{tcolorbox}
\textbf{RQ\textsubscript{1} (Seed Generation):}
VAEs' performance declined progressively as dataset complexity increased, resulting in only 14 correctly predicted seeds for ImageNet. In contrast, both GANs and DMs maintained consistently high accuracy in seed generation, with more than 66\% of seeds correctly predicted for all datasets.
\end{tcolorbox}

\subsection{RQ$_2$ (Effectiveness)}

The fifth column of \autoref{tab:comparison} presents the percentages and quantities of misclassification-inducing inputs, generated by GenAI TIGs, starting from the correctly predicted seeds (RQ\textsubscript{1}).

For low perturbation steps, DMs significantly outperformed or were comparable to other models (e.g., for the \textit{pizza} class), triggering up to 5$\times$ more failures for MNIST. Across all models, performance improved as dataset complexity increased. This is likely due to the growing difficulty of the tasks and the corresponding decline in classifier accuracy (\autoref{tab:classifiers}).

For high perturbation steps, all TIGs achieved high misclassification rates, with the VAE performing worst at 96.97\%. This indicates that larger perturbation steps are effective in triggering misclassifications, even within a constrained budget. 

\begin{tcolorbox}
\textbf{RQ\textsubscript{2} (Effectiveness):} DMs emerged as the most effective in generating misclassification-inducing inputs, especially at low perturbation steps, outperforming VAEs and GANs TIGs in most cases. Under high perturbation steps, all models showed considerable success in triggering misclassifications, indicating that increased perturbations uniformly promote misbehaviors across different architectures and datasets.
\end{tcolorbox}

\subsection{RQ$_3$ (Efficiency)}
The sixth column of \autoref{tab:comparison} reports the average number of iterations required by each TIG to trigger misclassifications.

For both low and high perturbation steps, DMs required significantly fewer iterations than other GenAI TIGs for MNIST, SVHN, and CIFAR-10. However, for ImageNet tasks, VAEs caused misclassifications in the fewest iterations.
This result reveals an interesting trade-off with VAEs: although they perform poorly in terms of the number of usable seeds and misclassification-inducing inputs (RQ\textsubscript{1} and RQ\textsubscript{2}), they compensate by requiring fewer iterations to trigger misclassifications.
Specifically, VAEs generate images with lower quality than more sophisticated models, particularly for complex datasets such as ImageNet, resulting in fewer acceptable test inputs. On the other hand, VAEs' training process tend to produce a smoother and more continuous latent space, which allows for more efficient exploration, i.e., changes in the latent vector tend to have a direct impact on the generated image.

\begin{tcolorbox}
\textbf{RQ\textsubscript{3} (Efficiency):} DMs were the most efficient for MNIST, SVHN, and CIFAR-10. In contrast, VAEs demonstrated the highest efficiency for ImageNet tasks. Across all GenAI TIGs, increasing the perturbation step consistently reduced the number of iterations needed to cause misclassifications, highlighting that larger perturbations accelerate the process of uncovering classifier weaknesses.
\end{tcolorbox}

\subsection{RQ$_4$ (Validity)}
The seventh column of \autoref{tab:comparison} shows the percentage and number of misclassification-inducing inputs deemed valid by human validators, excluding those where assessors disagreed.

For simpler datasets (MNIST and SVHN), VAEs either outperformed or performed comparably to other models at high perturbation steps. In these cases, DMs produced more valid inputs at low perturbation steps, despite showing lower percentages overall, which is partially due to their higher number of viable seeds (as discussed in RQ\textsubscript{1}).
For more complex datasets (CIFAR-10 and ImageNet), DMs generate significantly more valid inputs. For instance, DMs produced nearly 6$\times$ more valid misclassification-inducing inputs than VAEs for the \textit{teddy bear} class in ImageNet.

Interestingly, we observed that higher perturbation steps within the same GenAI TIG do not always compromise input validity. For instance, for the ImageNet \textit{teddy bear} class, both the number and percentage of valid inputs increased across all GenAI TIGs. This suggests that larger perturbations can still produce semantically valid inputs while being more effective at inducing misclassifications (as seen in RQ\textsubscript{3}).

\begin{tcolorbox}
\textbf{RQ\textsubscript{4} (Validity):}
According to humans, DMs excel at generating valid misclassification-inducing inputs for complex datasets like CIFAR-10 and ImageNet. For simpler datasets, GenAI TIGs demonstrate different trade-offs between the number and ratio of valid inputs, depending on the chosen perturbation step.
\end{tcolorbox}

\subsection{RQ$_5$ (Label Preservation)}
The eighth column of \autoref{tab:comparison} reports the percentage and number of preserved labels, which represent the valid misclassification-inducing inputs that maintained the ground-truth labels of their corresponding seeds. These inputs are particularly valuable for software testers.

GANs achieved remarkable results for simpler datasets (MNIST and SVHN), particularly at high perturbation steps, with up to 7$\times$ more label-preserving inputs than other GenAI models.
For more complex datasets like CIFAR-10 and ImageNet, GANs exhibited a performance decline, yielding the worst results, while DMs consistently outperformed the others, preserving up to 32$\times$ more labels than GANs for CIFAR-10 under high perturbation steps. DMs achieved a high percentage of preserved labels (i.e., $> 86\%$) for the ImageNet \textit{pizza} class, while nearly half of the labels were not preserved for the \textit{teddy bear} class, highlighting variability in label preservation across different tasks even within the same dataset.

\begin{tcolorbox}
\textbf{RQ\textsubscript{5} (Label Preservation):}
DMs achieve superior label preservation for complex datasets, achieving up to 92.86\% preserved labels. For simpler datasets, GANs frequently provided better or comparable performance than/to other GenAI TIGs.
\end{tcolorbox}

\subsection{Threats to Validity}

\head{Internal Validity} 
To mitigate threats due to uncontrolled variables, we integrated all GenAI models into a unified TIG framework, ensuring consistent parameters across all experiments, e.g., number of iterations and configuration of the selection operators. Moreover, we studied the potential impact of the perturbation step by performing experiments with low and high pertubation steps, obtained following a standardized procedure. Another possible threat may arise from survey participants providing unreliable answers. We addressed this threat by incorporating an ACQ in each survey, and limiting participation to workers with high reputations. 

\head{External Validity}
A possible threat is the choice of models and datasets. To mitigate it, we chose four popular datasets of increasing complexity, including ImageNet-1K, which contains high-resolution images from 1,000 different classes. We also considered widely recognized GenAI models from the literature, although our results may not generalize to all architectures. We aim to expand our comparison in the future.

\head{Reproducibility} We have made our experimental data and models publicly available~\cite{repo}.

%% file: table_for_compare_model.tex
\begin{table*}[t]
\caption{Comparison between GenAI TIGs across different datasets and mutation extents in terms of viable seeds, misclassification-inducing inputs, number of iterations to generate failure, input validity, and label preservation. The best results are highlighted in bold, while the underlined values are not statistically different from the best.}
\label{tab:comparison}
\begin{adjustbox}{width=2\columnwidth}
\renewcommand{\arraystretch}{1}
\begin{tabular}{lllrrrrrr}

\toprule                                                                  
\textbf{Dataset} 
& \textbf{Pert. Step ($\delta_{\textit{init}}$)} 
& \textbf{Model} 
& \textbf{\% Seeds} 
& \textbf{\% Misclass. (\#)} 
& \textbf{\# Iterations} 
& \textbf{\% Validity (\#)} 
& \textbf{\% Preserved (\#)} \\ 
 
\midrule

\multirow{6}{*}{MNIST} & \multirow{3}{*}{Low} & VAE & \textbf{99} & 4.04 (4) & 245.41 & 50.00 (2) & \textbf{100.00} (2) \\
& & GAN & \textbf{99} & 8.08 (8) & 242.05 & \textbf{75.00} (6) & 83.33 (5) \\
& & DM & 87 & \textbf{50.57} (\textbf{44}) & \textbf{164.61} & 45.45 (\textbf{20}) & 30.00 (\textbf{6}) \\ [0.5em]

& \multirow{3}{*}{High} & VAE & \textbf{99} & \textbf{100.00} (\textbf{99}) & 62.92 & \textbf{73.74 (73)} & \underline{49.32} (36) \\
& & GAN & \textbf{99} & 96.97 (96) & 107.46 & \underline{69.79} (67) & \textbf{62.69 (42)} \\
& & DM & 87 & \textbf{100.00} (87) & \textbf{26.77} & 40.23 (35) & 34.29 (12) \\ 

\midrule

\multirow{6}{*}{SVHN} & \multirow{3}{*}{Low} & VAE & 66 & 50.00 (33) & 178.20 & \textbf{51.52} (17) & 41.18 (7) \\
& & GAN & 84 & 42.86 (36) & 182.22 & 30.56 (11) & \underline{45.45} (5) \\
& & DM & \textbf{95} & \textbf{69.47} (\textbf{66}) & \textbf{131.04} & 39.39 (\textbf{26}) & \textbf{57.69} (\textbf{15}) \\ [0.5em]

& \multirow{3}{*}{High} & VAE & 66 & \textbf{100.00} (66) & 27.00 & \textbf{39.39} (26) & 30.77 (8) \\
& & GAN & 84 & \underline{98.81} (83) & 39.00 & \underline{36.14} (\textbf{30}) & \textbf{50.00} (\textbf{15}) \\
& & DM & \textbf{95} & \textbf{100.00} (\textbf{95}) & \textbf{13.23} & 23.16 (22) & 18.18 (4) \\ 

\midrule

\multirow{6}{*}{CIFAR-10} & \multirow{3}{*}{Low} & VAE & 39 & \underline{82.05} (32) & 118.90 & \underline{53.13} (17) & 29.41 (5) \\
& & GAN & 69 & 66.67 (46) & 140.32 & 45.65 (21) & 19.05 (4) \\
& & DM & \textbf{87} & \textbf{89.66} (\textbf{78}) & \textbf{85.63} & \textbf{60.26 (47)} & \textbf{61.70} (\textbf{29}) \\ [0.5em]

& \multirow{3}{*}{High} & VAE & 39 & \textbf{100.00} (39) & \underline{19.51} & 30.77 (12) & 33.33 (4) \\
& & GAN & 69 & \textbf{100.00} (69) & \underline{25.78} & 31.88 (22) & 22.73 (5) \\
& & DM & \textbf{87} & \textbf{100.00} (\textbf{87}) & \textbf{14.18} & \textbf{62.07 (54)} & \textbf{68.52} (\textbf{37}) \\ 

\midrule

\multirow{6}{*}{ImageNet (Teddy Bear)} & \multirow{3}{*}{Low} & VAE & 14 & \textbf{100.00} (14) & \textbf{13.57} & 78.57 (11) & \textbf{81.82} (9) \\
& & GAN & \underline{85} & \textbf{100.00} (85) & 98.27 & 74.12 (63) & 36.51 (23) \\
& & DM & \textbf{87} & \underline{98.85} (\textbf{86}) & 48.45 & \textbf{91.86} (\textbf{79}) & 49.37 (\textbf{39}) \\ [0.5em]

& \multirow{3}{*}{High} & VAE & 14 & \textbf{100.00} (14) & \textbf{1.36} & \textbf{100.00} (14) & \textbf{64.29} (9) \\
& & GAN &\underline{85} & \textbf{100.00} (85) & 26.38 & 83.53 (71) & 32.39 (23) \\
& & DM & \textbf{87} & \textbf{100.00} (\textbf{87}) & 6.63 & \underline{94.25} (\textbf{82}) & \underline{56.10} (\textbf{46}) \\ 

\midrule

\multirow{6}{*}{ImageNet (Pizza)} & \multirow{3}{*}{Low} & VAE & 25 & \textbf{100.00} (25) & \textbf{12.96} & \underline{92.00} (23) & \underline{91.30} (21) \\
& & GAN & \textbf{99} & 88.00 (87) & 172.88 & 88.51 (\textbf{77}) & 46.75 (36) \\
& & DM & 73 & \underline{97.26} (71) & 83.60 & \textbf{98.59} (70) & \textbf{92.86} (\textbf{65}) \\ [0.5em]

& \multirow{3}{*}{High} & VAE & 25 & \textbf{100.00} (25) & \textbf{2.60} & 80.00 (20) & \underline{75.00} (15) \\
& & GAN & \textbf{99} & \textbf{100.00} (99) & 47.93 & 86.87 (\textbf{86}) & 51.16 (44) \\
& & DM & 73 & \textbf{100.00} (73) & 12.53 & \textbf{100.00} (73) & \textbf{86.30} (\textbf{63}) \\ 

\bottomrule

\end{tabular}%
\end{adjustbox}
\end{table*}

%% file: 6-discussion.tex
\section{Lessons Learned and Key Insights}\label{sec:Discussion}

\subsection{Advanced GenAI Models Excel in Complex Tasks, but Their Superior Performance Comes at a Higher Cost}

VAEs and GANs performed well on less complex tasks, despite their simpler architectures and training processes compared to DMs. For instance, GANs produced the highest number of valid label-preserving misclassification-inducing inputs for MNIST and SVHN at high perturbation steps. Instead, DMs struggled with simpler datasets, as their lower resolution and limited variation do not offer enough complexity to fully exploit the diffusion process.
However, when considering more realistic datasets, such as CIFAR-10 and ImageNet, DMs clearly outperform other models, producing more viable seeds and up to 32$\times$ more label preserving inputs.

This superior performance comes with a trade-off in model cost. As emerges from \autoref{tab:gendl}, VAEs and GANs have much faster inference times than DMs, which is crucial for automated testing, especially for search-based TIGs that generate multiple inputs for several iterations.
DMs consistently take longer to generate an image compared to VAEs and GANs, mainly because the former architecture involves multiple steps of adding and removing noise. This process is time-consuming and demands more memory and processing resources, making it less efficient for tasks that require quick results.

For these reasons, testers should carefully assess task complexity and their available budget before beginning a testing campaign: in some cases, a simpler GenAI model may be more suitable than the latest, most advanced, architectures.

\subsection{Higher Perturbation Steps Speed Up Test Generation Without Compromising Input Validity or Label Preservation}

Latent space exploration, guided by our fitness function, effectively directed test generation to trigger misclassifications. As is common in search-based testing, increasing the strength of the mutation, i.e., the perturbation step, tends to improve efficiency. However, a frequent concern is that disruptive changes in inputs may compromise their validity and reduce their usefulness for testing~\cite{2023-Riccio-ICSE}.

In our experiments, higher perturbation steps consistently reduced the number of iterations needed to trigger failures. However, we found no clear evidence that increased perturbation steps negatively impacted input validity or label preservation across all GenAI models. 

One explanation for this phenomenon is the presence of mechanisms ensuring the adherence of the generative process to the target distribution, as well as the encoding of the desired input features (e.g., label conditions in conditional GANs or textual prompts in DMs). This capability highlights the potential of GenAI models for producing novel and meaningful test inputs.
Moreover, our TIG framework enforces the generation of latent vectors that remain within the observed distribution ranges reconstructed by GenAI models through the clamping operation. Clamping restricts the values of mutated latent vectors to a predefined range observed in the training data, mitigating the risk of exceeding boundaries that could lead to invalid inputs. Preliminary experiments confirmed that this mechanism is essential for maintaining input validity, as also highlighted by recent studies~\cite{dunn2021exposing,2023-Riccio-ICSE}.

\subsection{Latent Vectors Should Be Carefully Constrained and Carefully Manipulated}

Although constraining latent vectors was essential for effective test generation, our study identified limitations of GenAI TIGs. For example, even advanced TIGs generated at most 15 valid, label-preserving inputs out of 100 SVHN seeds. For the ImageNet \textit{pizza} class, nearly half of the valid inputs did not preserve the expected label.

Latent vectors constraints are less intuitive and interpretable than preconditions in traditional software, which are typically derived from domain-specific requirements.
Moreover, latent space exploration remains a challenging and open research area. Advancements like formulating constraints over the geometry of the latent space~\cite{dola2024cit4dnn}, latent space regularization~\cite{weiss2023generating}, or anomaly detection~\cite{stocco2020misbehaviour} should be generalized to more complex architectures and integrated into TIGs.

%% file: 7-related-work.tex
\section{Related Work}\label{sec:related-work}

In the literature, TIGs for DL based image classification have been largely compared on test effectiveness. These works define a certain TIG more effective than others if it can expose a higher number of misclassifications, regardless of the validity of such inputs.
Researchers also proposed and adopted adequacy criteria specific to DL systems, such as neuron-based coverage criteria~\cite{pei2017deepxplore, ma2018deepgauge, ma2019deepct}, which assess the extent to which test inputs exercise specific sets of neurons or DL model layers. Despite neuron-based coverage criteria have been extensively used to evaluate TIGs~\cite{tian2018deeptest, guo2018dlfuzz, demir2019deepsmartfuzzer, xie2019deephunter, dola2021distribution}, empirical results showed that higher neuron coverage may lead to the generation of invalid inputs~\cite{harel2020neuron}.
Other studies~\cite{riccio2021deepmetis} adopted mutant adequacy, e.g., the statistical notion of mutation adequacy introduced by Jahangirova and Tonella~\cite{jahangirova2020empirical}. These studies assess whether TIGs can expose DL model mutations, i.e., artificially injected faults that simulate real faults~\cite{humbatova2020taxonomy}.
The aforementioned works do not consider GenAI TIGs and mostly overlook the notion of test input validity and label preservation, which may influence their results.

On the other hand, more and more recent studies are considering the validity assessment of generated synthetic inputs~\cite{dola2021distribution, 2023-Riccio-ICSE, jiang2024validity, zhang2024enhancing}. Automation of validity assessment has been achieved by measuring the reconstruction error of VAEs~\cite{dola2021distribution, stocco2020misbehaviour, jiang2024validity}. However, such automated validation does not consider label preservation, primarily focusing on outlier detection as a proxy for validity.
Other research has involved human assessors in evaluating TIGs. Tian et al.~\cite{tian2021extent} demonstrated, through human assessment, that DL image classifier predictions are often unreliable, as they are influenced more by the surrounding context than by the predicted object. Attaoui et al.~\cite{attaoui2023black} involved industry practitioners to evaluate their feature extraction and clustering techniques for DL systems. 
Riccio et al. and Zhang et al. conducted studies on input validity, involving both automated validators and human assessors~\cite{2023-Riccio-ICSE, zohdinasab2023empirical, zhang2024enhancing}. They adopted a human evaluation similar to ours to perform a comprehensive comparison of different TIGs from the literature. 
Unlike their work, we consider a broader range of classification tasks and focus specifically on TIGs based on GenAI models, including the latest advancements, i.e., diffusion models. While their work only focus on validity and label preservation, we also consider effectiveness and efficiency, by providing a framework to fairly compare the impact of these models on test generation.

%% file: 8-conclusion.tex
\section{Conclusions and Future Work}\label{sec:conclusions}

In this paper, we present a comprehensive comparison of various GenAI models based on their ability to generate misclassification-inducing inputs for testing image classifiers. To achieve this, we introduced a search-based test generation framework that integrates different GenAI models and manipulates inputs by perturbing their latent space.

Our findings demonstrate that GenAI models can successfully generate valid, label-preserving and failure-inducing inputs across all considered classification tasks. Notably, simpler GenAI models perform well on less complex tasks such as MNIST and SVHN, while more advanced models are needed for more challenging ones, i.e., CIFAR-10 and ImageNet. Additionally, we found that increasing the latent vector perturbation step accelerates test generation without compromising input validity or label preservation.

This study and the proposed framework open up several avenues for future research. We plan to conduct a broader evaluation with more datasets and GenAI architectures and extend our framework to incorporate more sophisticated search algorithms and additional testing objectives, such as input diversity and mutation killing.

%% file: paper.bbl
\begin{thebibliography}{10}
\providecommand{\url}[1]{#1}
\csname url@samestyle\endcsname
\providecommand{\newblock}{\relax}
\providecommand{\bibinfo}[2]{#2}
\providecommand{\BIBentrySTDinterwordspacing}{\spaceskip=0pt\relax}
\providecommand{\BIBentryALTinterwordstretchfactor}{4}
\providecommand{\BIBentryALTinterwordspacing}{\spaceskip=\fontdimen2\font plus
\BIBentryALTinterwordstretchfactor\fontdimen3\font minus
  \fontdimen4\font\relax}
\providecommand{\BIBforeignlanguage}[2]{{%
\expandafter\ifx\csname l@#1\endcsname\relax
\typeout{** WARNING: IEEEtran.bst: No hyphenation pattern has been}%
\typeout{** loaded for the language `#1'. Using the pattern for}%
\typeout{** the default language instead.}%
\else
\language=\csname l@#1\endcsname
\fi
#2}}
\providecommand{\BIBdecl}{\relax}
\BIBdecl

\bibitem{wang2021comparative}
P.~Wang, E.~Fan, and P.~Wang, ``{Comparative analysis of image classification
  algorithms based on traditional machine learning and deep learning},''
  \emph{Pattern Recognition Letters}, vol. 141, pp. 61--67, 2021.

\bibitem{sarwinda2021deep}
D.~Sarwinda, R.~H. Paradisa, A.~Bustamam, and P.~Anggia, ``{Deep Learning in
  Image Classification using Residual Network (ResNet) Variants for Detection
  of Colorectal Cancer},'' \emph{Procedia Computer Science}, vol. 179, pp.
  423--431, 2021.

\bibitem{lou2022testing}
G.~Lou, Y.~Deng, X.~Zheng, M.~Zhang, and T.~Zhang, ``Testing of autonomous
  driving systems: where are we and where should we go?'' in \emph{Proceedings
  of the 30th ACM Joint European Software Engineering Conference and Symposium
  on the Foundations of Software Engineering}, 2022, pp. 31--43.

\bibitem{tang2023survey}
S.~Tang, Z.~Zhang, Y.~Zhang, J.~Zhou, Y.~Guo, S.~Liu, S.~Guo, Y.-F. Li, L.~Ma,
  Y.~Xue \emph{et~al.}, ``A survey on automated driving system testing:
  Landscapes and trends,'' \emph{ACM Transactions on Software Engineering and
  Methodology}, vol.~32, no.~5, pp. 1--62, 2023.

\bibitem{neelofar2024identifying}
N.~Neelofar and A.~Aleti, ``Identifying and explaining safety-critical
  scenarios for autonomous vehicles via key features,'' \emph{ACM Transactions
  on Software Engineering and Methodology}, vol.~33, no.~4, 2024.

\bibitem{guerriero2021operation}
A.~Guerriero, R.~Pietrantuono, and S.~Russo, ``{Operation is the hardest
  teacher: estimating DNN accuracy looking for mispredictions},'' in
  \emph{Proceedings of the 43rd International Conference on Software
  Engineering}.\hskip 1em plus 0.5em minus 0.4em\relax IEEE Press, 2021, p.
  348–358.

\bibitem{zohdinasab2023deepatash}
T.~Zohdinasab, V.~Riccio, and P.~Tonella, ``{DeepAtash: Focused Test Generation
  for Deep Learning Systems},'' in \emph{Proceedings of the 32nd ACM SIGSOFT
  International Symposium on Software Testing and Analysis}.\hskip 1em plus
  0.5em minus 0.4em\relax New York, NY, USA: Association for Computing
  Machinery, 2023, p. 954–966.

\bibitem{zhangTSE22}
J.~M. Zhang, M.~Harman, L.~Ma, and Y.~Liu, ``Machine learning testing: Survey,
  landscapes and horizons,'' \emph{IEEE Transactions on Software Engineering},
  vol.~48, no.~1, pp. 1--36, 2022.

\bibitem{RiccioEMSE20}
V.~Riccio, G.~Jahangirova, A.~Stocco, N.~Humbatova, M.~Weiss, and P.~Tonella,
  ``Testing machine learning based systems: a systematic mapping,''
  \emph{Empirical Software Engineering}, vol.~25, 2020.

\bibitem{braiek2020testing}
H.~B. Braiek and F.~Khomh, ``On testing machine learning programs,''
  \emph{Journal of Systems and Software}, vol. 164, 2020.

\bibitem{2023-Riccio-ICSE}
V.~Riccio and P.~Tonella, ``When and why test generators for deep learning
  produce invalid inputs: an empirical study,'' in \emph{2023 IEEE/ACM 45th
  International Conference on Software Engineering (ICSE)}, 2023.

\bibitem{baresi2001test}
L.~Baresi and M.~Young, ``Test oracles,'' \emph{Technical Report CIS-TR-01-02},
  2001.

\bibitem{pezze2014automated}
M.~Pezze and C.~Zhang, ``Automated test oracles: A survey,'' in \emph{Advances
  in computers}.\hskip 1em plus 0.5em minus 0.4em\relax Elsevier, 2014,
  vol.~95, pp. 1--48.

\bibitem{jahangirova2016test}
G.~Jahangirova, D.~Clark, M.~Harman, and P.~Tonella, ``{Test oracle assessment
  and improvement},'' in \emph{Proceedings of the 25th International Symposium
  on Software Testing and Analysis}.\hskip 1em plus 0.5em minus 0.4em\relax New
  York, NY, USA: Association for Computing Machinery, 2016.

\bibitem{dola2021distribution}
S.~Dola, M.~B. Dwyer, and M.~L. Soffa, ``Distribution-aware testing of neural
  networks using generative models,'' in \emph{2021 IEEE/ACM 43rd International
  Conference on Software Engineering (ICSE)}.\hskip 1em plus 0.5em minus
  0.4em\relax IEEE, 2021.

\bibitem{pei2017deepxplore}
K.~Pei, Y.~Cao, J.~Yang, and S.~Jana, ``Deepxplore: Automated whitebox testing
  of deep learning systems,'' in \emph{proceedings of the 26th Symposium on
  Operating Systems Principles}, 2017, pp. 1--18.

\bibitem{guo2018dlfuzz}
J.~Guo, Y.~Jiang, Y.~Zhao, Q.~Chen, and J.~Sun, ``Dlfuzz: Differential fuzzing
  testing of deep learning systems,'' in \emph{Proceedings of the 2018 26th ACM
  Joint Meeting on European Software Engineering Conference and Symposium on
  the Foundations of Software Engineering}, 2018.

\bibitem{ma2018deepgauge}
L.~Ma, F.~Juefei-Xu, F.~Zhang, J.~Sun, M.~Xue, B.~Li, C.~Chen, T.~Su, L.~Li,
  Y.~Liu \emph{et~al.}, ``Deepgauge: Multi-granularity testing criteria for
  deep learning systems,'' in \emph{Proceedings of the 33rd ACM/IEEE
  international conference on automated software engineering}, 2018.

\bibitem{braiek2019deepevolution}
H.~B. Braiek and F.~Khomh, ``Deepevolution: A search-based testing approach for
  deep neural networks,'' in \emph{2019 IEEE International Conference on
  Software Maintenance and Evolution (ICSME)}.\hskip 1em plus 0.5em minus
  0.4em\relax IEEE, 2019.

\bibitem{riccio2020model}
V.~Riccio and P.~Tonella, ``Model-based exploration of the frontier of
  behaviours for deep learning system testing,'' in \emph{Proceedings of the
  28th ACM Joint Meeting on European Software Engineering Conference and
  Symposium on the Foundations of Software Engineering}, 2020.

\bibitem{zohdinasab2021deephyperion}
T.~Zohdinasab, V.~Riccio, A.~Gambi, and P.~Tonella, ``Deephyperion: exploring
  the feature space of deep learning-based systems through illumination
  search,'' in \emph{Proceedings of the 30th ACM SIGSOFT International
  Symposium on Software Testing and Analysis}, 2021, pp. 79--90.

\bibitem{Abdessalem-ICSE18}
R.~{Ben Abdessalem}, S.~{Nejati}, L.~{C. Briand}, and T.~{Stifter}, ``Testing
  vision-based control systems using learnable evolutionary algorithms,'' in
  \emph{2018 IEEE/ACM 40th International Conference on Software Engineering
  (ICSE)}, May 2018, pp. 1016--1026.

\bibitem{fahmy2021supporting}
H.~Fahmy, F.~Pastore, M.~Bagherzadeh, and L.~Briand, ``Supporting deep neural
  network safety analysis and retraining through heatmap-based unsupervised
  learning,'' \emph{IEEE Transactions on Reliability}, vol.~70, no.~4, pp.
  1641--1657, 2021.

\bibitem{kang2020sinvad}
S.~Kang, R.~Feldt, and S.~Yoo, ``Sinvad: Search-based image space navigation
  for dnn image classifier test input generation,'' in \emph{Proceedings of the
  IEEE/ACM 42nd International Conference on Software Engineering Workshops},
  2020, pp. 521--528.

\bibitem{dunn2021exposing}
I.~Dunn, H.~Pouget, D.~Kroening, and T.~Melham, ``Exposing previously
  undetectable faults in deep neural networks,'' in \emph{Proceedings of the
  30th ACM SIGSOFT International Symposium on Software Testing and Analysis},
  2021, pp. 56--66.

\bibitem{aleti2023software}
A.~Aleti, ``Software testing of generative ai systems: Challenges and
  opportunities,'' in \emph{2023 IEEE/ACM International Conference on Software
  Engineering: Future of Software Engineering (ICSE-FoSE)}, 2023.

\bibitem{dola2024cit4dnn}
S.~Dola, R.~McDaniel, M.~B. Dwyer, and M.~L. Soffa, ``Cit4dnn: Generating
  diverse and rare inputs for neural networks using latent space combinatorial
  testing,'' in \emph{Proceedings of the IEEE/ACM 46th International Conference
  on Software Engineering}, 2024, pp. 1--13.

\bibitem{goodfellow2020generative}
I.~Goodfellow, J.~Pouget-Abadie, M.~Mirza, B.~Xu, D.~Warde-Farley, S.~Ozair,
  A.~Courville, and Y.~Bengio, ``Generative adversarial networks,''
  \emph{Communications of the ACM}, vol.~63, no.~11, 2020.

\bibitem{kang2024deceiving}
S.~Kang, R.~Feldt, and S.~Yoo, ``Deceiving humans and machines alike:
  Search-based test input generation for dnns using variational autoencoders,''
  \emph{ACM Transactions on Software Engineering and Methodology}, vol.~33,
  no.~4, pp. 1--24, 2024.

\bibitem{repo}
Maryam, M.~Biagiola, A.~Stocco, and V.~Riccio, ``Replication package,''
  \url{https://github.com/Maryammaryam877/genai_tigs}, 2024.

\bibitem{kingma2013auto}
D.~P. Kingma, ``Auto-encoding variational bayes,'' \emph{arXiv preprint
  arXiv:1312.6114}, 2013.

\bibitem{10.1214/aoms/1177729694}
S.~Kullback and R.~A. Leibler, ``{On Information and Sufficiency},'' \emph{The
  Annals of Mathematical Statistics}, vol.~22, no.~1, 1951.

\bibitem{sohl2015deep}
J.~Sohl-Dickstein, E.~Weiss, N.~Maheswaranathan, and S.~Ganguli, ``Deep
  unsupervised learning using nonequilibrium thermodynamics,'' in
  \emph{International conference on machine learning}.\hskip 1em plus 0.5em
  minus 0.4em\relax PMLR, 2015.

\bibitem{ho2020denoising}
J.~Ho, A.~Jain, and P.~Abbeel, ``Denoising diffusion probabilistic models,''
  \emph{Advances in neural information processing systems}, vol.~33, 2020.

\bibitem{Abdessalem-ASE18-1}
R.~B. Abdessalem, A.~Panichella, S.~Nejati, L.~C. Briand, and T.~Stifter,
  ``Testing autonomous cars for feature interaction failures using
  many-objective search,'' in \emph{Proceedings of ASE '18}, ser. ASE
  2018.\hskip 1em plus 0.5em minus 0.4em\relax New York, NY, USA: ACM, 2018.

\bibitem{Abdessalem-ASE18-2}
R.~{Ben Abdessalem}, S.~{Nejati}, L.~C. {Briand}, and T.~{Stifter}, ``Testing
  advanced driver assistance systems using multi-objective search and neural
  networks,'' in \emph{31st IEEE/ACM International Conference on Automated
  Software Engineering (ASE)}, 2016.

\bibitem{gambi2019automatically}
A.~Gambi, M.~Mueller, and G.~Fraser, ``Automatically testing self-driving cars
  with search-based procedural content generation,'' in \emph{Proceedings of
  the 28th ACM SIGSOFT International Symposium on Software Testing and
  Analysis}, 2019, pp. 318--328.

\bibitem{SD_latent_walk}
I.~Stenbit, F.~Chollet, and W.~Luke, ``A walk through latent space with stable
  diffusion,''
  \url{https://keras.io/examples/generative/random_walks_with_stable_diffusion/},
  2023, accessed: 10-09-2024.

\bibitem{nichol2022glidephotorealisticimagegeneration}
A.~Nichol, P.~Dhariwal, A.~Ramesh, P.~Shyam, P.~Mishkin, B.~McGrew,
  I.~Sutskever, and M.~Chen, ``Glide: Towards photorealistic image generation
  and editing with text-guided diffusion models,'' 2022.

\bibitem{getimg_ai_guide}
\BIBentryALTinterwordspacing
Getimg.ai, ``Interactive guide to stable diffusion guidance scale parameter,''
  accessed: 1 January 2024. [Online]. Available:
  \url{https://getimg.ai/guides/interactive-guide-to-stable-diffusion-guidance-scale-parameter}
\BIBentrySTDinterwordspacing

\bibitem{pmlr-v70-guo17a}
C.~Guo, G.~Pleiss, Y.~Sun, and K.~Q. Weinberger, ``On calibration of modern
  neural networks,'' in \emph{Proceedings of the 34th International Conference
  on Machine Learning}, vol.~70.\hskip 1em plus 0.5em minus 0.4em\relax PMLR,
  2017.

\bibitem{zhang2024enhancing}
J.~Zhang, J.~Keung, X.~Ma, X.~Li, Y.~Xiao, Y.~Li, and W.~K. Chan, ``Enhancing
  valid test input generation with distribution awareness for deep neural
  networks,'' in \emph{2024 IEEE 48th Annual Computers, Software, and
  Applications Conference (COMPSAC)}.\hskip 1em plus 0.5em minus 0.4em\relax
  IEEE, 2024, pp. 1095--1100.

\bibitem{pytorch-vgg-source}
{PyTorch Contributors}, ``Source code for torchvision.models.vgg,''
  \url{https://pytorch.org/vision/0.12/\_modules/torchvision/models/vgg.html\#vgg19},
  2024, accessed: 2024-01-01.

\bibitem{LecunBBH98}
Y.~Lecun, L.~Bottou, Y.~Bengio, and P.~Haffner, ``{Gradient-based learning
  applied to document recognition},'' \emph{Proceedings of the IEEE}, vol.~86,
  no.~11, pp. 2278--2324, 1998.

\bibitem{Netzer2011}
Y.~Netzer, T.~Wang, A.~Coates, A.~Bissacco, B.~Wu, A.~Y. Ng \emph{et~al.},
  ``{Reading digits in natural images with unsupervised feature learning},'' in
  \emph{NIPS Workshop on Deep Learning and Unsupervised Feature Learning},
  2011.

\bibitem{Krizhevsky2009}
\BIBentryALTinterwordspacing
A.~Krizhevsky, ``Learning multiple layers of features from tiny images,'' in
  \emph{Master's thesis, University of Toronto}, 2009. [Online]. Available:
  \url{https://api.semanticscholar.org/CorpusID:18268744}
\BIBentrySTDinterwordspacing

\bibitem{Deng2009}
J.~Deng, W.~Dong, R.~Socher, L.-J. Li, K.~Li, and L.~Fei-Fei, ``{ImageNet: A
  large-scale hierarchical image database},'' in \emph{2009 IEEE Conference on
  Computer Vision and Pattern Recognition}.\hskip 1em plus 0.5em minus
  0.4em\relax IEEE, 2009.

\bibitem{tensorflow-keras-example}
TensorFlow, ``Keras example,''
  \url{https://www.tensorflow.org/datasets/keras_example}, accessed:
  2022-08-29.

\bibitem{kang2020VAEMNISTcode}
S.~Kang, R.~Feldt, and S.~Yoo, ``Vae training code for mnist,''
  \url{https://github.com/coinse/SINVAD/blob/master/vae/train.py}, 2020,
  accessed: August 1, 2023.

\bibitem{pytorch_dcganCode}
PyTorch, ``Dcgan example in pytorch,''
  \url{https://github.com/pytorch/examples/blob/main/dcgan/main.py}, accessed:
  September 2, 2023.

\bibitem{pytorch_cdcganarchitecture}
Zokovie, ``Cdcgan-mnist,'' \url{https://github.com/zokovi/cDCGAN-MNIST},
  accessed: September 3, 2023.

\bibitem{kohya_ss_lora_gui_codefile}
B.~Maltais, ``Kohya ss - lora gui code,''
  \url{https://github.com/bmaltais/kohya_ss/blob/master/kohya_gui/lora_gui.py},
  accessed: November 30, 2023.

\bibitem{sinvad2020VAEConvCodearchitecture}
S.~Kang, R.~Feldt, and S.~Yoo, ``Vae convolutional training code,''
  \url{https://github.com/coinse/SINVAD/blob/master/vae/train_conv.py},
  accessed: September 29, 2023.

\bibitem{maunish2023VAEImagenetkaggle}
M.~Bhandari, ``Training vae on imagenet with pytorch,''
  \url{https://www.kaggle.com/code/maunish/training-vae-on-imagenet-pytorch},
  accessed: April 2024.

\bibitem{brock2018large}
A.~Brock, J.~Donahue, and K.~Simonyan, ``Large scale {GAN} training for high
  fidelity natural image synthesis,'' in \emph{International Conference on
  Learning Representations}, 2019.

\bibitem{luo2021case}
J.~Luo, J.~Huang, and H.~Li, ``A case study of conditional deep convolutional
  generative adversarial networks in machine fault diagnosis,'' \emph{Journal
  of Intelligent Manufacturing}, vol.~32, no.~2, pp. 407--425, 2021.

\bibitem{stable-diffusion-v15}
``Stable diffusion v1.5 repository,''
  \url{https://huggingface.co/stable-diffusion-v1-5/stable-diffusion-v1-5},
  accessed: 2023-12-15.

\bibitem{hu2021lora}
E.~J. Hu, Y.~Shen, P.~Wallis, Z.~Allen-Zhu, Y.~Li, S.~Wang, L.~Wang, and
  W.~Chen, ``Lora: Low-rank adaptation of large language models,'' 2021.

\bibitem{stocco2020misbehaviour}
A.~Stocco, M.~Weiss, M.~Calzana, and P.~Tonella, ``Misbehaviour prediction for
  autonomous driving systems,'' in \emph{Proceedings of the ACM/IEEE 42nd
  international conference on software engineering}, 2020.

\bibitem{mturk}
Amazon, ``Mechanical turk,'' \url{https://www.mturk.com}.

\bibitem{behrend2011viability}
T.~S. Behrend, D.~J. Sharek, A.~W. Meade, and E.~N. Wiebe, ``The viability of
  crowdsourcing for survey research,'' \emph{Behavior research methods},
  vol.~43, pp. 800--813, 2011.

\bibitem{KitturCS08}
A.~Kittur, E.~H. Chi, and B.~Suh, ``Crowdsourcing user studies with mechanical
  turk,'' in \emph{Proceedings of the SIGCHI Conference on Human Factors in
  Computing Systems}.\hskip 1em plus 0.5em minus 0.4em\relax New York, NY, USA:
  ACM, 2008.

\bibitem{HeerB10}
J.~Heer and M.~Bostock, ``Crowdsourcing graphical perception: Using mechanical
  turk to assess visualization design,'' in \emph{Proceedings of the SIGCHI
  Conference on Human Factors in Computing Systems}, ser. CHI '10.\hskip 1em
  plus 0.5em minus 0.4em\relax New York, NY, USA: ACM, 2010.

\bibitem{PastoreMF13}
F.~{Pastore}, L.~{Mariani}, and G.~{Fraser}, ``Crowdoracles: Can the crowd
  solve the oracle problem?'' in \emph{2013 IEEE Sixth International Conference
  on Software Testing, Verification and Validation}, March 2013, pp. 342--351.

\bibitem{peer2014reputation}
E.~Peer, J.~Vosgerau, and A.~Acquisti, ``Reputation as a sufficient condition
  for data quality on amazon mechanical turk,'' \emph{Behavior research
  methods}, vol.~46, pp. 1023--1031, 2014.

\bibitem{Crowdsourcing6}
A.~W. Edwards, ``Ra fischer, statistical methods for research workers,
  (1925),'' in \emph{Landmark writings in western mathematics 1640-1940}.\hskip
  1em plus 0.5em minus 0.4em\relax Elsevier, 2005, pp. 856--870.

\bibitem{Wilcoxon1945}
F.~Wilcoxon, ``Individual comparisons by ranking methods,'' \emph{Biometrics
  Bulletin}, vol.~1, no.~6, 1945.

\bibitem{cohen1988statistical}
J.~Cohen, \emph{Statistical power analysis for the behavioral sciences}.\hskip
  1em plus 0.5em minus 0.4em\relax Hillsdale, N.J: L. Erlbaum Associates, 1988.

\bibitem{weiss2023generating}
M.~Weiss, A.~G. G{\'o}mez, and P.~Tonella, ``Generating and detecting true
  ambiguity: a forgotten danger in dnn supervision testing,'' \emph{Empirical
  Software Engineering}, vol.~28, no.~6, p. 146, 2023.

\bibitem{ma2019deepct}
L.~Ma, F.~Juefei-Xu, M.~Xue, B.~Li, L.~Li, Y.~Liu, and J.~Zhao, ``Deepct:
  Tomographic combinatorial testing for deep learning systems,'' in \emph{2019
  IEEE 26th International Conference on Software Analysis, Evolution and
  Reengineering (SANER)}.\hskip 1em plus 0.5em minus 0.4em\relax IEEE, 2019,
  pp. 614--618.

\bibitem{tian2018deeptest}
Y.~Tian, K.~Pei, S.~Jana, and B.~Ray, ``Deeptest: Automated testing of
  deep-neural-network-driven autonomous cars,'' in \emph{Proceedings of the
  40th international conference on software engineering}, 2018.

\bibitem{demir2019deepsmartfuzzer}
S.~Demir, H.~F. Eniser, and A.~Sen, ``Deepsmartfuzzer: Reward guided test
  generation for deep learning,'' \emph{arXiv preprint arXiv:1911.10621}, 2019.

\bibitem{xie2019deephunter}
X.~Xie, L.~Ma, F.~Juefei-Xu, M.~Xue, H.~Chen, Y.~Liu, J.~Zhao, B.~Li, J.~Yin,
  and S.~See, ``Deephunter: a coverage-guided fuzz testing framework for deep
  neural networks,'' in \emph{Proceedings of the 28th ACM SIGSOFT international
  symposium on software testing and analysis}, 2019, pp. 146--157.

\bibitem{harel2020neuron}
F.~Harel-Canada, L.~Wang, M.~A. Gulzar, Q.~Gu, and M.~Kim, ``Is neuron coverage
  a meaningful measure for testing deep neural networks?'' in \emph{Proceedings
  of the 28th ACM Joint Meeting on European Software Engineering Conference and
  Symposium on the Foundations of Software Engineering}, 2020, pp. 851--862.

\bibitem{riccio2021deepmetis}
V.~Riccio, N.~Humbatova, G.~Jahangirova, and P.~Tonella, ``Deepmetis:
  Augmenting a deep learning test set to increase its mutation score,'' in
  \emph{2021 36th IEEE/ACM International Conference on Automated Software
  Engineering (ASE)}.\hskip 1em plus 0.5em minus 0.4em\relax IEEE, 2021, pp.
  355--367.

\bibitem{jahangirova2020empirical}
G.~Jahangirova and P.~Tonella, ``An empirical evaluation of mutation operators
  for deep learning systems,'' in \emph{2020 IEEE 13th International Conference
  on Software Testing, Validation and Verification (ICST)}.\hskip 1em plus
  0.5em minus 0.4em\relax IEEE, 2020, pp. 74--84.

\bibitem{humbatova2020taxonomy}
N.~Humbatova, G.~Jahangirova, G.~Bavota, V.~Riccio, A.~Stocco, and P.~Tonella,
  ``Taxonomy of real faults in deep learning systems,'' in \emph{Proceedings of
  the ACM/IEEE 42nd international conference on software engineering}, 2020,
  pp. 1110--1121.

\bibitem{jiang2024validity}
Z.~Jiang, H.~Li, R.~Wang, X.~Tian, C.~Liang, F.~Yan, J.~Zhang, and Z.~Liu,
  ``Validity matters: Uncertainty-guided testing of deep neural networks,''
  \emph{Software Testing, Verification and Reliability}, p. e1894, 2024.

\bibitem{tian2021extent}
Y.~Tian, S.~Ma, M.~Wen, Y.~Liu, S.-C. Cheung, and X.~Zhang, ``To what extent do
  dnn-based image classification models make unreliable inferences?''
  \emph{Empirical Software Engineering}, vol.~26, no.~5, p.~84, 2021.

\bibitem{attaoui2023black}
M.~Attaoui, H.~Fahmy, F.~Pastore, and L.~Briand, ``Black-box safety analysis
  and retraining of dnns based on feature extraction and clustering,''
  \emph{ACM Transactions on Software Engineering and Methodology}, vol.~32,
  no.~3, 2023.

\bibitem{zohdinasab2023empirical}
T.~Zohdinasab, V.~Riccio, and P.~Tonella, ``An empirical study on low-and
  high-level explanations of deep learning misbehaviours,'' in \emph{2023
  ACM/IEEE International Symposium on Empirical Software Engineering and
  Measurement (ESEM)}.\hskip 1em plus 0.5em minus 0.4em\relax IEEE, 2023, pp.
  1--11.

\end{thebibliography}
